# A Comparative Study on Transfer Learning and Distance Metrics in Semantic Clustering over the COVID-19 Tweets

Elnaz Zafarani-Moattar[1], Mohammad Reza Kangavari [*, 2], Amir Masoud Rahmani [1]

[1] Department of Computer Engineering, Science and Research Branch, Islamic Azad University, Tehran, Iran
[2] Department of Computer Engineering, Iran University of Science and Technology, Tehran, Iran

[*] Corresponding Author: kangavari@iust.ac.ir

## Abstract

This paper is a comparison study in the context of Topic Detection on COVID-19 data. There are various approaches for Topic Detection, among which the Clustering approach is selected in this paper. Clustering requires distance and calculating distance needs embedding. The aim of this research is to simultaneously study the three factors of embedding methods, distance metrics and clustering methods and their interaction. A dataset including one-month tweets collected with COVID-19-related hashtags is used for this study. Five methods, from earlier to new methods, are selected among the embedding methods: Word2Vec, fastText, GloVe, BERT and T5. Five clustering methods are investigated in this paper that are: k-means, DBSCAN, OPTICS, spectral and Jarvis-Patrick. Euclidian distance and Cosine distance as the most important distance metrics in this field are also examined. First, more than 7,500 tests are performed to tune the parameters. Then, all the different combinations of embedding methods with distance metrics and clustering methods are investigated by silhouette metric. The number of these combinations is 50 cases. First, the results of these 50 tests are examined. Then, the rank of each method is taken into account in all the tests of that method. Finally, the major variables of the research (embedding methods, distance metrics and clustering methods) are studied separately. Averaging is performed over the control variables to neutralize their effect. The experimental results show that T5 strongly outperforms other embedding methods in terms of silhouette metric. In terms of distance metrics, cosine distance is weakly better. DBSCAN is also superior to other methods in terms of clustering methods.

# 1- Introduction

In mid-December 2019, a Corona-type virus called COVID-19 was reported in Wuhan, China, which quickly became the most important news headline worldwide and had wide reactions on social media. When this disease has been reported for the first time, no one could imagine that it would become a global epidemic and be continued in this way, and it would lead to severe damage to the economies of nations by widespread closures. The number of people infected by the coronavirus has reached more than 140 million in the world in April 2021, and the number of its victims has exceeded 3 million[1]. The United States is at the first rank having more than 32 million COVID-19 patients [2], followed by India, Brazil and France.

The high importance of the COVID-19 epidemic and its worldwide prevalence has led to the daily production of large volumes of messages on the topic of COVID-19 on social media. The extent of the COVID-19 topic has also led to the production of these messages in many subtopics; for example, symptoms, the way to prevent, diagnose and treat, vaccine production process, as well as its impact on the economy, individual life, and people's mental health (e.g. mental health of people has been studied in [3], [4]). Each of these subtopics is considered as a topic with a high volume of published messages. There are different classifications regarding the methods used for topic/event detection [5]–[9]. In 2018, Rania Ibrahim et al. [7] classified topic detection methods into five classes: Clustering, Frequent Pattern Mining, Exemplar-Based, Matrix Factorization and Probabilistic models. This paper is placed in the clustering class.

The primary aim of this paper is to study and compare the clustering methods used in topic detection. However, before clustering, embedding methods are needed to convert input texts to vectors. Then it is necessary to measure the distance between these embedding vectors. Therefore, the existing distance metrics should be used. So, a comparative approach is applied here. As a result, the goal of this paper is to study and compare the performance of possible combinations among different selected clustering algorithms, embedding methods and distance metrics. The tweets posted on the Twitter social network on the topic of COVID-19 is used as a dataset. Twitter is chosen because it is very popular and used by most scientific articles in this field. According to the above description, three major questions are considered for this paper:

1) Which of the embedding methods has better performance in topic detection on COVID-19 tweets?

2) Which of the distance metrics has better performance in topic detection on COVID-19 tweets?
3) Which of the clustering methods has better performance in topic detection on COVID-19 tweets?

## 2- Literature Review

There are many articles in the field of COVID-19 disease, including three fields of image, dataset and text. This work is in the field of text. Most of the previous works on COVID-19 are in the field of image, which have been performed on X-ray images of the chest[10]–[12]. In fact, they attempt to detect healthy people from COVID-19 patients by the lung image using deep learning methods. In the field of data, various datasets have been compiled, such as Cord19 and Cord19STS datasets. There are several works in the field of text, such as sentiment analysis and topic detection (topic modeling). In the following, each field will be studied in detail.

Since all NLP based works require a dataset, the topic of providing a dataset has been investigated from different aspects. In CORD-19 [13], 128,000 articles with keywords such as COVID, Coronavirus and 2019-nCoV have been collected from publishers such as Elsevier, Springer, etc. After collecting, clustering and duplicates removal have been applied to the articles. In CORD19STS [14], a parameter called STS[1] is measured on CORD-19 articles and manually annotated by labels such as Related, Somewhat-related and Not-related by AMT[2] users. In [15], a corpus with 7500 tweets about the corona test has been provided and manually labeled in five classes of events with topics of Tested Positive, Tested Negative, Can Not Test, Death and Cure & Prevention. For each of these events, a number of questions are addressed, like who, when, and where has a positive or negative corona test? In [16], a multi-language dataset including 6 million tweets has been collected. Considering the distribution of collected tweets shows that English has the highest rate among the 66 languages in the dataset, and 63% of the tweets are in English. Also, in [17], a dataset containing 123 million tweets has been published, in which 60% of tweets are in English. In [18], a dataset called CovidQA has been presented, which contains 124 question-article pairs. In [19], a dataset called TweetsCOV19 has been introduced, which contains more than 11 million tweets. Metadata about tweets such as entities, hashtags, user mentions, sentiments and URLs are also extracted. There is also a dataset on Twitter [20] in the sentiment analysis area, which classifies tweets into five classes as follows: positive, very positive, neutral, negative and very negative. Also, 10 topics are specified, and the relation of each tweet with these 10 topics are determined. However, this dataset includes tweets that do not belong to any of these topics or belong to several topics simultaneously. In [21], the COV19Tweets dataset is introduced, which has more than 310 million tweets in English along with a sentiment score. The Geo version is presented as GeoCOV19Tweets dataset, which includes tweets originated from 204 different countries, and the United States has the largest rate (43%).

[1] Semantic Textual Similarity
[2] Amazon Mechanical Turk

The works in the field of text are mostly about topic detection/modeling and sentiment analysis. Some examples will be discussed in the following paragraphs.

One of the works in the field of sentiment analysis is presented in [22]. Four classification algorithms are considered in this work: Linear Regression Model, Naïve Bayes Classifier, Logistic Regression and K-Nearest Neighbor. Here, the length of the tweet is also taken into account to evaluate the performance of each method in short tweets and longer tweets. Textual data visualization is also used to identify the critical trend of change in fear-sentiment. In [23], topic detection and sentiment analysis are studied on the Reddit social network. The research framework consists of four steps. In step 1, COVID-19-related comments are collected from the Reddit social network. Step 2 includes preprocessing of data. In step 3, the LDA method is used for topic detection. In step 4, the combination of embedding and LSTM is used for sentiment classification on COVID-19 comment. The GloVe 50-dimensional is used for embedding. In [24], eleven salient topics with the highest score in the LDA method are selected. In addition to these topics, bigram[3] is also specified. Eight emotions, such as Anger, Fear, Joy, Sadness, etc., are considered. The experimental results show that the feeling of fear is prominent. In [25] and [26], LDA method is used for topic modeling, and VADER[4] is applied for sentiment analysis. VADER is a lexicon and rule-based sentiment analysis tool that is specifically attuned to the sentiment expressed in social media. In [27], a neural network for sentiment analysis using multilingual sentence embedding is presented. This network is trained on the Sentiment140 dataset. Also, it is tested on pre-trained word- and sentence-level embedding models: Word2Vec, BERT, ELMO and MUSE[5]. In addition, tree map of Twitter activity in Europe during the time period from December 2019 to April 2020 is plotted. The United Kingdom, Spain, Germany, Italy, and France have the highest contribution of tweets.

[3] The most popular pairs of words within each topic
[4] Valence Aware Dictionary and sEntiment Reasoner
[5] Multilingual Universal Sentence Encoder

A sample of text classification works is presented in [28], which uses KNN, LR[6] and SVM for classification. In classification, topics must be predefined as labels in advance. In this paper, 11 labels are defined as follows: Donate, News & Press, Prevention, Reporting, Share, Speculation, Symptoms, Transmission, Travel, Treatment, What Is Corona?. The following sample of text classification is [29], which is done on Sina Weibo social network. Here, Weibo data are classified into seven types of situations, using supervised learning methods, namely SVM, Naïve Bayes and Random Forest.

# 3- Materials and Methods

As mentioned in the introduction section, a comparative approach is selected to compare between three components: embedding methods, distance metrics and clustering methods. The proposed framework will be presented in section 4, and it will be mentioned that the main processing phase of the framework includes these three components. Therefore, these three components are described in this section.

## 3-1- Embedding Methods

The main concept of word embedding is that any words in a language can be expressed by a vector of numbers. Usually, the existing word embedding models have vectors with the length between 200 and 750. Word embedding consists of n-dimensional vectors which try to record the meaning of words and their content by numerical values. Each set of numbers can be considered as a "word vector", but it is not necessarily useful. A useful word embedding vector represents the meaning of that word, and the similar words reach similar embedding values. In other words, semantic similarity leads in vector similarity. Therefore, using these vectors for clustering results in semantic clustering.

In this paper, five word embedding methods are selected and investigated, namely, Word2Vec, fastText, GloVe, BERT and T5. Word2Vec is one of the earliest embedding methods, and T5 is one of the latest and most up-to-date embedding methods. We will briefly review each method in the following.

### Word2vec (2013)

Word2Vec has been developed in 2013 by a Mikolov et al. at Google [30], Although the one-hot encoding existed before Word2Vec, Word2Vec is the first embedding method in practice; because in one-hot coding, we cannot make any significant comparison between two

[6] logistic regression

vectors and can only check whether the vectors are equal. In other words, semantic similarity does not lead to vector similarity in one-hot encoding. However, in Word2Vec, the vectors are calculated in a way that the semantic similarity leads to vector similarity. In Word2Vec, a distributed representation is used for each word. It means that each word is represented by a distribution of numerical values (weights) on different elements of the vector. Therefore, the representation of a word is distributed across all the elements of a vector instead of one-to-one mapping between an element in a vector and a word in a dictionary and each element contribute to the meaning of a large number of words. As a result, different words are described by different values of each element, and all elements contribute to recording the meaning of words.

### fastText (2016)

fastText is a library created in 2016 by Facebook's AI Research (FAIR) lab [31], which is applied for learning word embedding and text classification [32]. The fastText uses the vector representation combination of word sections to represent a word vector. The models which assign a distinct vector to each word do not consider the word's morphology. This representation is considered as a constraint for languages that have a large number of words where there are so many rare words. The applied method in fastText is based on skip-gram model, where each word is represented as a bag of character n-grams, and a vector is assigned to each character n-grams. The sum of these vectors forms the word representation. In other words, the representation of each word is obtained from the sum of the character n-gram vectors of that word. This method has a high speed and is trained quickly on large corpora. This model is also able to create a word representation vector for words not appeared in the training dataset.

### GloVe (2014)

GloVe is the abbreviation for Global Vector. This model was developed in 2014 at Stanford University as an open-source project [33]. This model is an unsupervised learning algorithm to obtain vector representation for words. This is achieved by mapping the words within a meaningful space where the distance of the words' vector is related to their semantic similarity. The log-bilinear regression model is used for unsupervised learning of word representation. This model combines the advantages and features of global matrix factorization and local context window models. This model acts based on statistical information and trains only non-zero elements in the word-by-word co-occurrence matrix instead of the entire sparse matrix. Although the methods such as LSA, as a member of the global matrix factorization family, are efficient in statistical information, they have weak performance in word analogy. In contrast, methods such as skip-gram, from the local context window family, perform better in terms of word analogy, but perform poorly in terms of corpus statistics, as they use the local window in training instead of global co-occurrence.

#### Bert Embedding (2018)

In 2018, a big model called BERT [34] was trained by Google engineers with lots of data (Wikipedia + Book Corpus) and made available for NLP issues. In 2017, Vaswani et al. at Google had published a paper entitled "Attention Is All You Need" [35] and introduced the concept of transformer neural network, which has been used in BERT. Two methods can be used to train BERT: Masked Language Model and Next Sentence Prediction. The given model can be used in two ways: feature extraction and fine-tuning. The BERT model is trained in two different sizes: The base BERT consists of 12 Encoder layers (called Transformer Blocks in the original article), and the larger network consists of 24 Encoder layers.

#### T5 Embedding (2020)

In 2020, Colin Raffel et al. has introduced their proposed framework, called T5: "Text-to-Text Transfer Transformer" [36]. The idea behind T5 is to consider each text processing problem as a Text to Text problem, i.e. taking the text as an input and generating the new text as an output. In the T5 architecture, the original Transformer architecture of Vaswani et al. [37] is used, but there are three differences: First, the Norm bias layer is removed. Secondly, the layer normalization is out of the residual path. Third, a different position embedding scheme is applied.

T5 is commonly used as pretrained. The pretrained T5 model is publicly available. Unlike machine vision, where the network is supervisely-pretrained on labeled data, in NLP, unsupervised learning is often used on unlabeled data. Therefore, a large-scale dataset, called C4, is used to train T5. C4, which stands for "Colossal Clean Crawled Corpus", has been gathered and introduced for T5 training for the first time. This dataset has been built as a source for unlabeled text, which includes hundreds of gigabytes of clean English text.

## 3-2- Similarity Metrics

Metrics are needed to measure the degree of similarity or distance among texts in text clustering. As we said, the texts will be converted to embedding vectors. Since Euclidean distance and cosine similarity are the most commonly used metrics in the vector space, they are used in this research. Suppose A = $[a_1, a_2, \ldots, a_n]$ and B = $[b_1, b_2, \ldots, b_n]$ are two vectors (points) in the n-dimensional Euclidean space. Their Euclidean distance is defined as:

$$d(A,B) = \sqrt{\sum_{i=1}^{n}(a_i - b_i)^2} \tag{1}$$

Since this distance is not normal, the normalized Euclidean distance is used, which is obtained as:

$$Dis_{Euc}(A,B) = \frac{d(A,B)}{\max_{i,j}\{d(A_i,B_j)\}} \quad (2)$$

If it is needed, the Euclidean distance is converted to Euclidean similarity using the following formula:

$$Sim_{\text{Euc}}(A,B) = 1 - Dis_{\text{Euc}}(A,B) \quad (3)$$

The cosine similarity between two vectors A and B is also calculated as:

$$Sim_{\text{Cos}}(A,B) = \frac{A.B}{\|A\|\|B\|} = \frac{\sum_{i=1}^{n} a_i b_i}{\sqrt{\sum_{i=1}^{n} {a_i}^2}\sqrt{\sum_{i=1}^{n} {b_i}^2}} \quad (4)$$

The cosine similarity is converted to cosine distance using the following formula whenever it is needed:

$$Dis_{\text{Cos}}(A,B) = 1 - Sim_{\text{Cos}}(A,B) \quad (5)$$

### 3-3- Clustering Methods

The works on Topic Detection follow two different approaches: classification and clustering. In the classification approach, the number and type of classes are predetermined. This approach is suitable for applications in which classes are well known. For example, news agencies classify news into predefined classes such as economics, health, sports, and so on. However, in the field of social media, a new topic may occur at any moment, and the previous topics will fade over time. Therefore, classification algorithms will not be applicable for this task and clustering algorithms must be used.

In this paper, the common clustering algorithms used in topic detection are compared. k-means, DBSCAN [38], OPTICS [39], spectral [40] and Jarvis-Patrick [41] clustering algorithms are selected for this purpose. Among these methods, k-means is placed in the class of partitioning methods; DBSCAN and OPTICS are density-based methods; and spectral and Jarvis-Patrick are graph-based methods.

## 4- System Framework

The flowchart of our work is shown in Figure 1. As shown in the figure, the proposed framework consists of three phases: preprocessing, main processing, and postprocessing. Tweet stream fed into the preprocessing phase, which includes two components: First, text preprocessing is applied to each tweet's text. Then the tweets are filtered, which means a decision is made over keeping or removing the tweet. The results of this phase are stored temporarily. Then the messages are entered into the main processing phase as a chunk of tweets, which includes three components: embedding, distance measuring and clustering. Finally, in the third phase, postprocessing and visualization are applied to the results.

## 4-1- Preprocessing

The preprocessing phase has two components: text preprocessing and tweet preprocessing. Text preprocessing consists of the following steps: the input text is tokenized, a tag is added for tokens which are numbers or punctuation; also, the language tag (English word, Persian word, etc.), tag for emoji, hashtag, URL, mention and other special characters are added; then the remaining text is converted to lowercase, and stop words are removed. A special tokenizer developed in our lab (ComInSys lab) is used for this purpose which is able to perform all these steps except lower case conversion and stop word removal. Tweet preprocessing is the next component in which tweets in other languages and tweets with less than four remaining words are removed.

## 4-2- Main Processing

The main processing phase consists of three components: embedding, distance measuring and clustering. First of all, the tweets are embedded by one of the given methods (i.e. each tweet is embedded by each embedding method). Then the distance between every two obtained vectors is measured by distance metrics (i.e. the distance between each vector and any other vector is measured by both Euclidean distance and cosine distance). Next, the embedding vectors of tweets, along with the distance of each pair, are given to each clustering method in order to extract clusters of tweets. In other words, each tweet is embedded by each embedding method, then the distance of each pair is measured by each distance metric, and finally, all tweets are clustered by each clustering method. As a result, the combination of these components makes 5×2×5=50 cases. Therefore, 50 tests have been performed for these 50 cases, and the results are given in the next section.

Among the embedding methods, Word2Vec, fastText, GloVe, BERT and T5 are used. The embedding vector dimension for each method is given in Table 1. Word2Vec is fine-tuned over COVID data, fastText and GloVe are fine-tuned over tweeter data, and basic versions of BERT and T5 are used. Therefore, transfer learning is used in this paper. Transfer learning in deep learning and specially NLP, enabled researchers to use pretrained models for various problems [42].

Table 1: Dimension of embedded vectors for each embedding method.

| Embedding method | Word2Vec | fastText | GloVe | BERT | T5 |
|---|---|---|---|---|---|
| Dimension of embedded vector | 400 | 400 | 200 | 768 | 768 |

## 4-3- Postprocessing

The postprocessing phase has two components: postprocessing of clusters and visualization of the results. The postprocessing of clusters consists of topic extraction and calculating silhouette measure. The Silhouette measure lets us compare different combinations of the components discussed in the previous section. The main advantage of silhouette measure for our work is that it can be used without ground truth knowledge. Word cloud is one of the visualization results of the system. A sample of a word cloud produced by the system is shown in Figure 2.

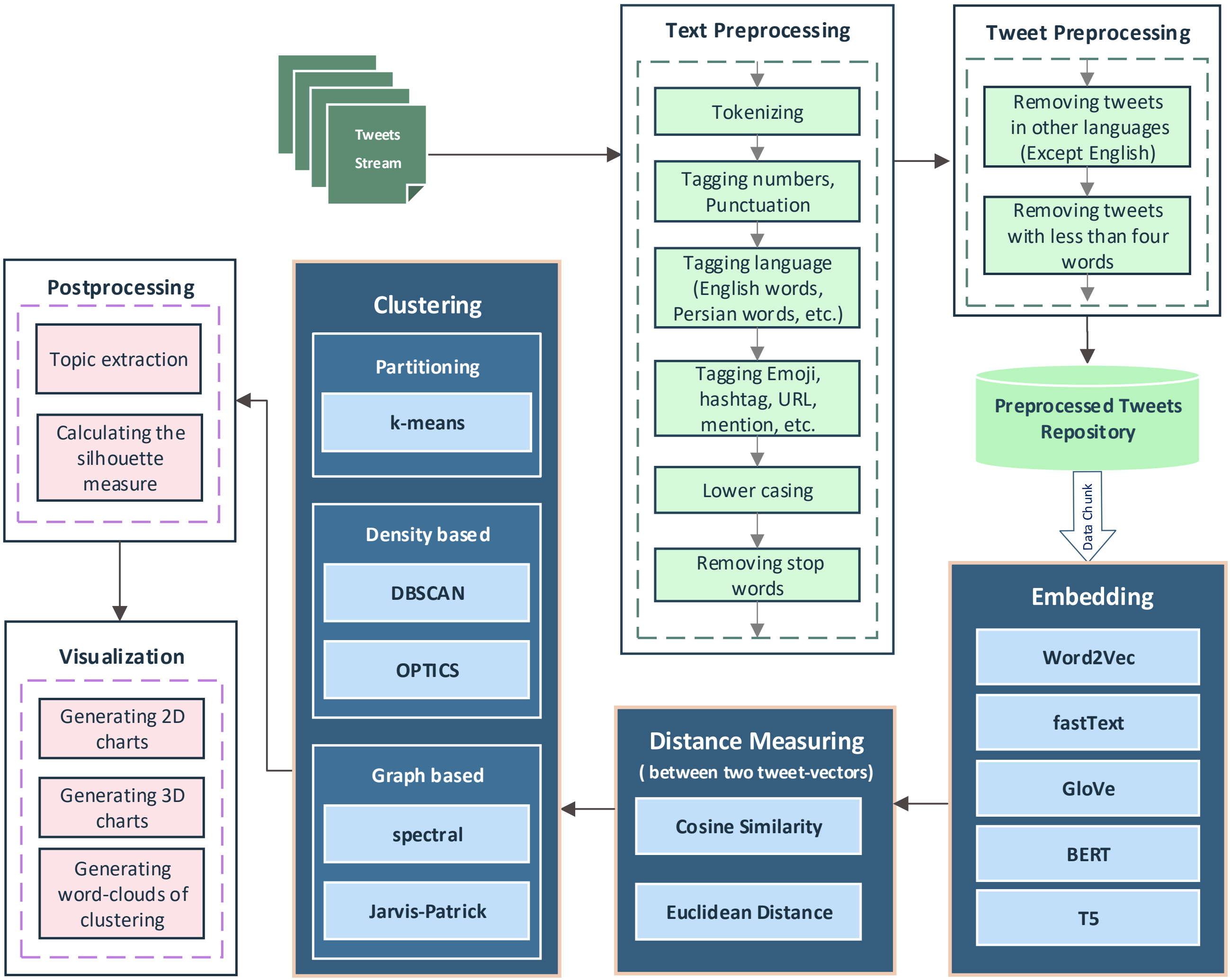

Figure 1: Overview of the research framework

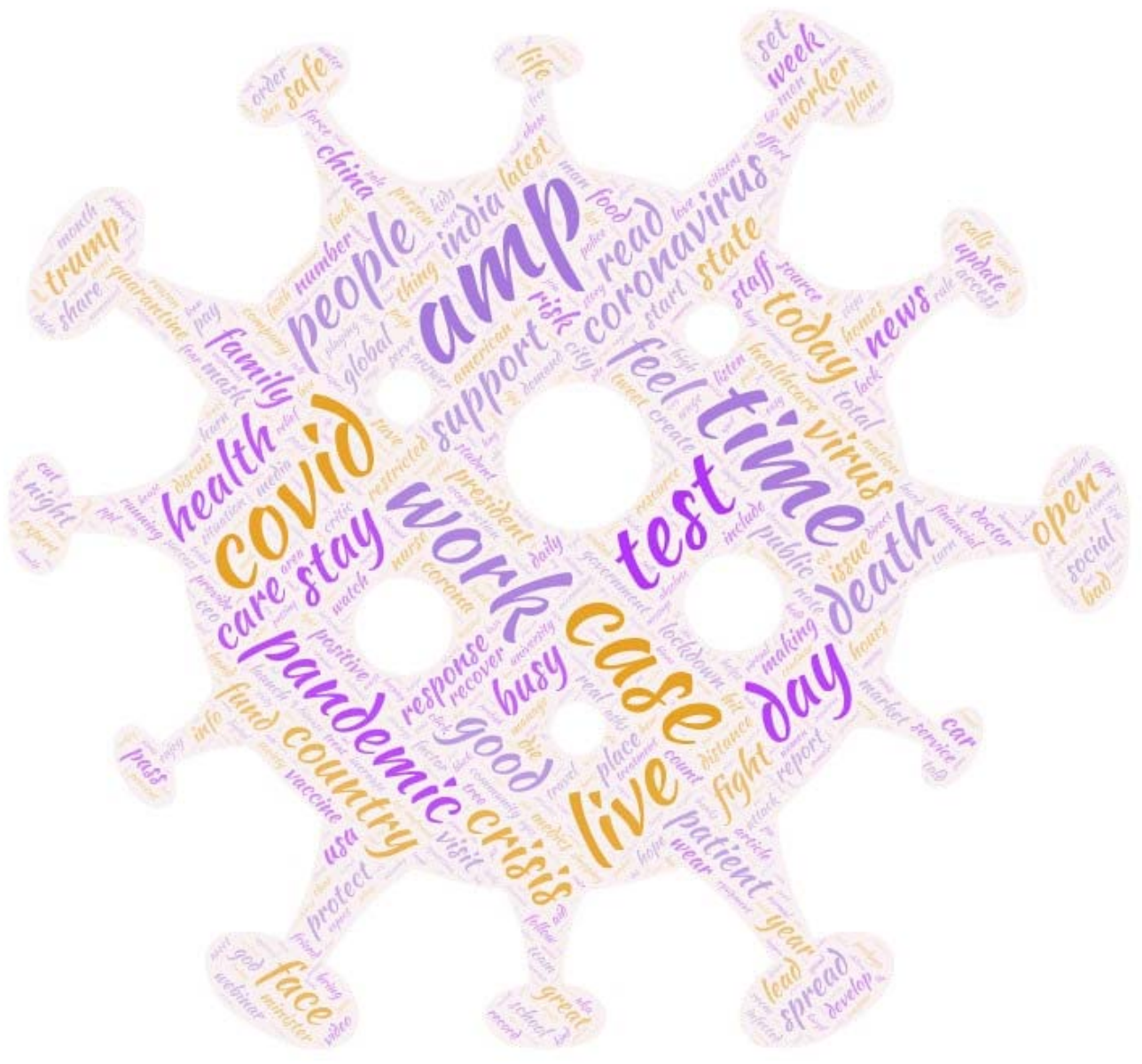

Figure 2: A sample of word cloud produced by the system

## 5- Implementation Details

In this section, the details of implementation are considered. First, the dataset and then the evaluation metric are presented. Finally, the details of each experiment and its results are discussed.

### 5-1- Data Set

This research's dataset includes tweets from March 29, 2020, to August 30, 2020, taken from [43], [44]. This dataset contains tweets from users who have used the following hashtags: #coronavirus, #coronavirusoutbreak, #coronavirusPandemic, #covid19, #covid_19, #epitwitter, #ihavecorona, #StayHomeStaySafe, #TestTraceIsolate. The collected dataset has multiple fields such as: the context of different tweets, tweeted account, used hashtags, accounts location, tweet language, and retweets argument. As the processing has been done in English, English-related tweets are taken from the dataset (i.e. tweets whose "Lang" field is "En"). In addition, the retweets argument is set to FALSE. Therefore, the dataset does not contain retweets, but the number of retweets is considered as a variable.

### 5-2- Evaluation Metric

Silhouette measure is used to evaluate performance and compare alternative methods in different experiments. The main advantage of silhouette measure for our work is that it can be used without any ground truth knowledge.

For each data point i in the cluster $C_m$, silhouette value s(i) is calculated as:

$$s(i) = \begin{cases} 0 & if\ |C_m| = 1 \\ \frac{b(i)-a(i)}{max\{a(i),b(i)\}} & if\,|C_m| > 1 \end{cases} \quad (6)$$

in which a(i) and b(i) are calculated as:

$$a(i) = \frac{1}{|C_m|-1}\sum_{j\in C_m,\ j\neq i} d(i,j) \quad (7)$$

$$b(i) = \min_{n\neq m} \frac{1}{|C_n|}\sum_{j\in C_n} d(i,j) \quad (8)$$

where d(i, j) is the distance between data points i and j.

## 5-3- Experiments and Results

In this section, the experiments to answer the research questions are considered. Since the applied clustering methods have parameters to be set, it is necessary to perform experiments to set these parameters at the first stage. Table 2 lists the parameters of different methods along with the range of values assigned in the experiments. As it can be seen, some methods have a single parameter, and others have two parameters. For methods with two parameters, both two parameters have been tuned simultaneously. Figure 3 and Figure 4 show the results of some experiments as a sample. The optimal parameter value for single-parameter methods (k-means, OPTICS and spectral) and two-parameter methods (DBSCAN and Jarvis-Patrick) are selected by eq. (9) and eq. (10), respectively.

$$p^* = \underset{p}{Argmax}\{Exp(i,j,k,p)\}\ \forall p\epsilon Domain(p), i\epsilon Emb, j\epsilon Dis, k\epsilon Cls \quad (9)$$

$$\langle p_1{}^*, p_2{}^*\rangle = \underset{\langle p_1, p_2\rangle}{Argmax}\{Exp(i,j,k,p_1,p_2)\}\ \forall p_1\epsilon Domain(p_1) \wedge$$
$$\forall p_2\epsilon Domain(p_2), i\epsilon Emb, j\epsilon Dis, k\epsilon Cls \quad (10)$$

where $Exp(i,j,k,p)$ represents the performed experiment by embedder i, distance metric j and clustering method k using p-value for the given parameter (similarly, in $Exp(i,j,k,p_1,p_2)$, using $p_1$ and $p_2$ values). Then, the experiments with the optimal value of the parameters have been used for the next steps, i.e. the results are obtained by eq. (11) and (12):

$$Res(i,j,k) = Exp(i,j,k,p^*) \quad (11)$$

$$Res(i,j,k) = Exp(i,j,k,p_1^*,p_2^*) \quad (12)$$

Table 2: The applied clustering methods, their parameters, range of their values and the number of experiments which is performed during parameter tuning.

| Clustering Method | k-means | DBSCAN | | OPTICS | spectral | Jarvis-Patrick | |
|---|---|---|---|---|---|---|---|
| Parameter | k | ε | Min Pts | Min Pts | K | k | $k_t$ |
| Value Range | 2-49 | 0.1-5 | 2-49 | 2-49 | 2-49 | 10-100 | 1-k |
| Number of Experiments | 48 | 2400 | | 48 | 48 | 5005 | |

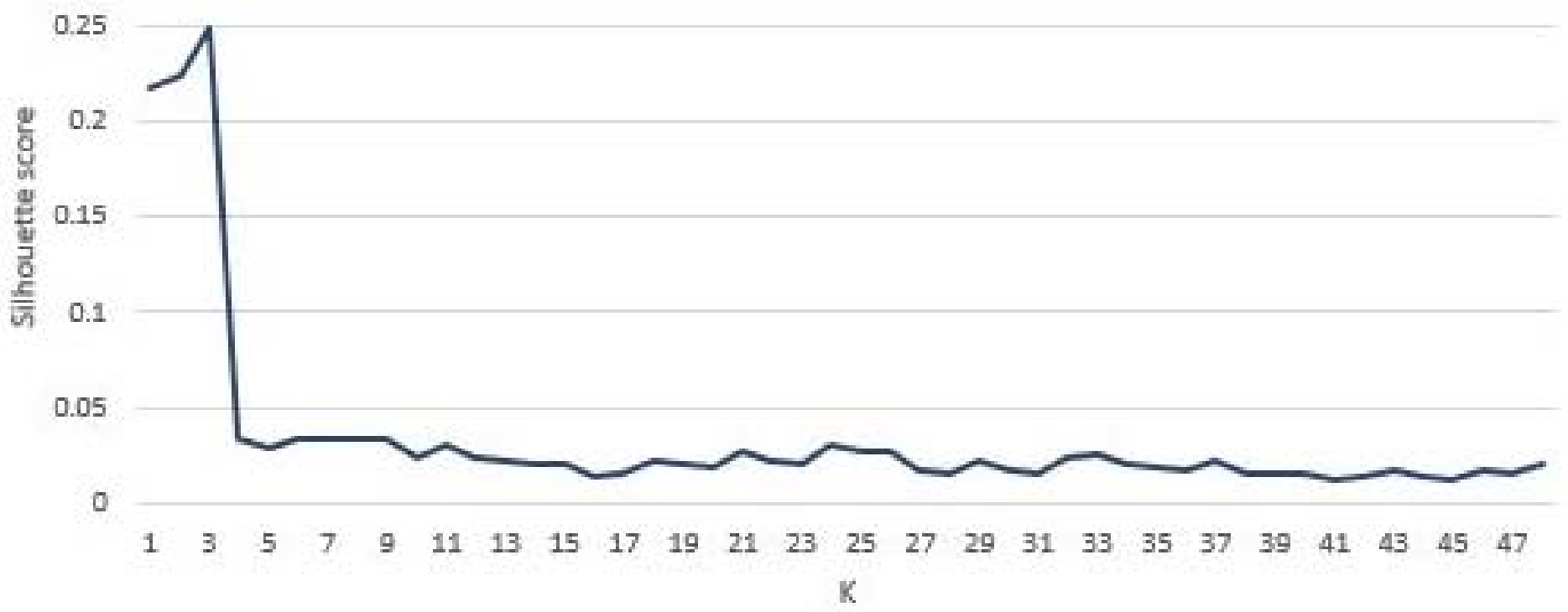

Figure 3: Spectral clustering performance measured by silhouette score for different values of k (measured on Word2Vec embedder and Euclidian distance)

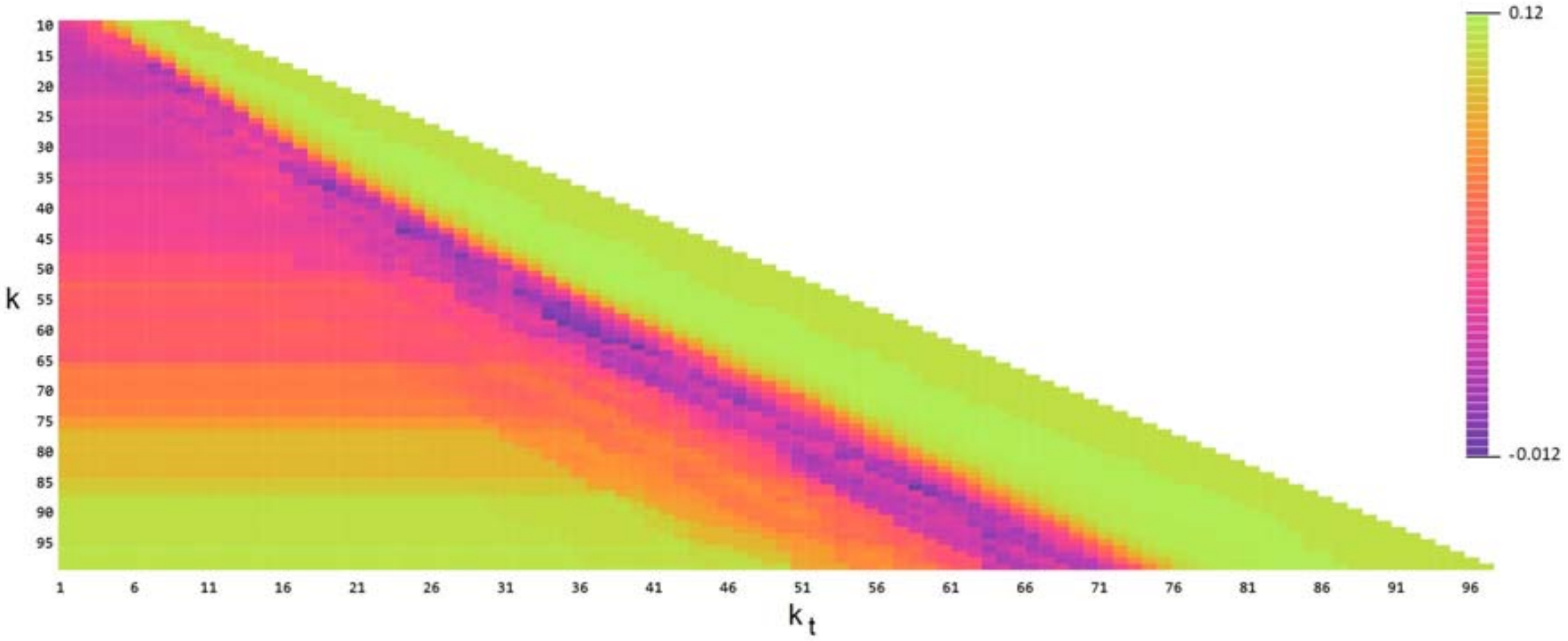

Figure 4: Heat map visualization of Jarvis-Patrick clustering performance on silhouette score. Each row corresponds to values of k, and each column corresponds to values of $k_t$. fastText embedder and cosine distance are used.

After tuning the parameters, the main experiments are performed to answer the main research questions. The experimental results are presented in Table 3. “Which embedding method is more effective in topic detection?” is the first question of this research. It is necessary to neutralize the effect of other variables in order to be able to compare different embedding methods. For this purpose, the final result of each embedding method is obtained by eq. (13).

$$R_E(i) = (|Dis|.|Cls|)^{-1} \sum_{j \epsilon Dis} \sum_{k \epsilon Cls} Res(i,j,k) \quad (13)$$

where $R_E(i)$ represents the final result of the $i^{th}$ embedding method. Figure 5 shows the obtained results. It can be seen that the T5 embedder has achieved the best result while the fastText fails to obtain good results. Word2Vec, BERT and GloVe have gained medium results. Table 3 shows that T5 has more stable results, as in most cases (6 of 10) it has the first rank. To verify it, the ranks of embedding methods are listed in

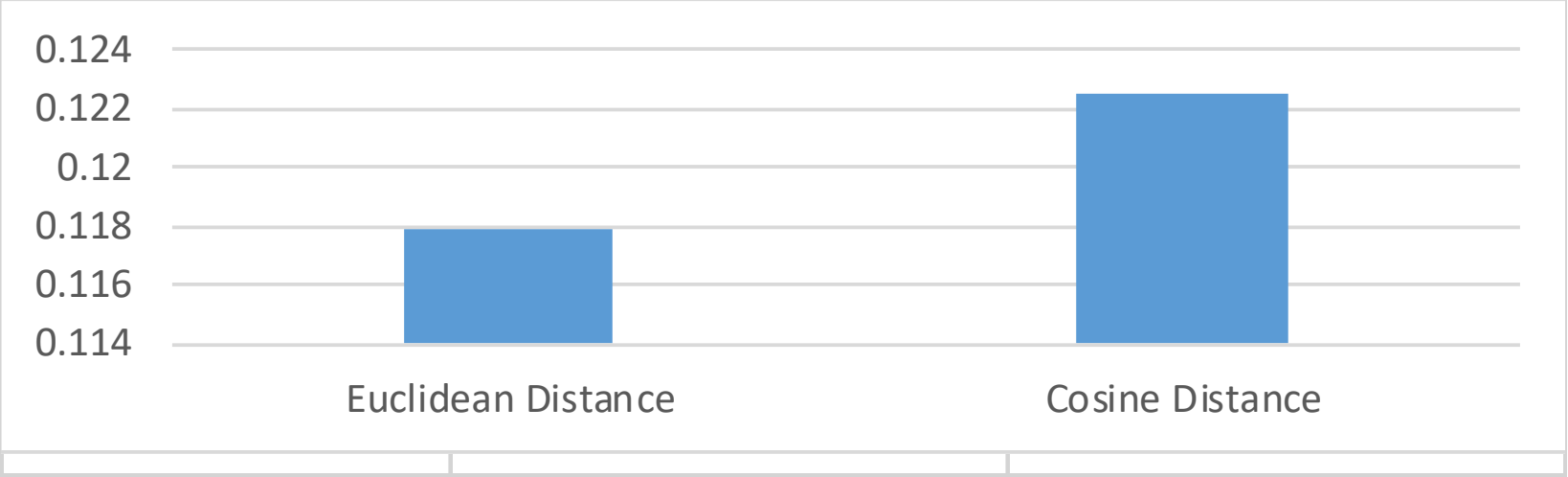

Figure 6: Final results for different distance metrics ($R_D$) obtained by eq. (6). (Averaged silhouette score.)

Table 4.

The second research question is: "which distance metric has better performance?". Same as the previous question, the final result of each distance metric is obtained by eq. (14):

$$R_D(j) = (|Emb|.|Cls|)^{-1} \sum_{i \epsilon Emb} \sum_{k \epsilon Cls} Res(i,j,k) \quad (14)$$

where $R_D(j)$ represents the final result of the $j^{th}$ distance metric. The results are shown in Figure 6. As it can be seen, the cosine distance is superior to the Euclidean distance. However, since its superiority is weak, we decide to perform more investigations. Therefore, we study the superiority of each one in the results of different embedding and clustering methods. The results are given in Table 5. According to the results, it is obvious that the superiority of cosine distance over Euclidean distance is weak (15 out of 25 cases). It is worth noting that when using k-means, cosine distance is a better metric than Euclidean distance. Also, this is almost true for Jarvis-Patrick.

Which clustering method outperforms others?" This is the third research question. Same as other questions, it is necessary to obtain the final result for each clustering method at the first stage. It has been done by eq. (15):

$$R_C(k) = (|Emb|.|Dis|)^{-1} \sum_{i \epsilon Emb} \sum_{j \epsilon Dis} Res(i,j,k) \quad (15)$$

where $R_c(k)$ represents the final result of the $k^{th}$ clustering method. Figure 7 shows the obtained results. As it can be seen, DBSCAN has better results over the other clustering methods, followed by spectral. Studying the partial results in different embedding and distance metric methods also indicates that in some cases, spectral has better results than DBSCAN. However, in general, DBSCAN is superior to spectral.

Table 3: Experimental results for different embedding and clustering methods and different distance metrics measured by silhouette score (Res(i, j, k))

| Clustering Method | Distance Metric | Embedding Method | Silhouette Coefficient |
| --- | --- | --- | --- |
| k-means | Euclidian | Word2Vec | 0/0359385 |
| | | GloVe | 0/0692724 |
| | | fastText | 0/0328571 |
| | | BERT | 0/0393112 |
| | | T5 | 0/1878792 |
| DBSCAN | Euclidian | Word2Vec | 0/32647333 |
| | | GloVe | 0/335792233 |
| | | fastText | 0/38503435 |
| | | BERT | -0/008437181 |
| | | T5 | 0/42997536 |
| OPTICS | Euclidian | Word2Vec | 0/24897604 |
| | | GloVe | -0/043140313 |
| | | fastText | -0/04975994 |
| | | BERT | -0/008437181 |
| | | T5 | -0/045669287 |
| spectral | Euclidian | Word2Vec | 0/24897604 |
| | | GloVe | 0/039996269 |
| | | fastText | 0/024566475 |
| | | BERT | 0/31211817 |
| | | T5 | 0/16695139 |
| Jarvis-Patrick | Euclidian | Word2Vec | 0/012290187 |
| | | GloVe | 0/010123885 |
| | | fastText | 0/009033865 |
| | | BERT | 0/035866346 |
| | | T5 | 0/150094329 |
| k-means | Cosine | Word2Vec | 0/06861783 |
| | | GloVe | 0/108646679 |
| | | fastText | 0/086493306 |
| | | BERT | 0/06051539 |
| | | T5 | 0/23428407 |
| DBSCAN | Cosine | Word2Vec | 0/10450104 |
| | | GloVe | 0/508365662 |
| | | fastText | 0/3217964 |
| | | BERT | 0/35840496 |
| | | T5 | 0 |
| OPTICS | Cosine | Word2Vec | 0/06944838 |
| | | GloVe | 0/042236351 |
| | | fastText | -0/16361202 |
| | | BERT | -0/06467958 |
| | | T5 | 0/009819393 |
| spectral | Cosine | Word2Vec | 0/10450104 |
| | | GloVe | 0/020867131 |
| | | fastText | 0/01713565 |
| | | BERT | 0/37993073 |
| | | T5 | 0/38004518 |
| Jarvis-Patrick | Cosine | Word2Vec | 0/086882312 |
| | | GloVe | 0/037956569 |
| | | fastText | 0/012601484 |
| | | BERT | 0/129499881 |
| | | T5 | 0/148158701 |

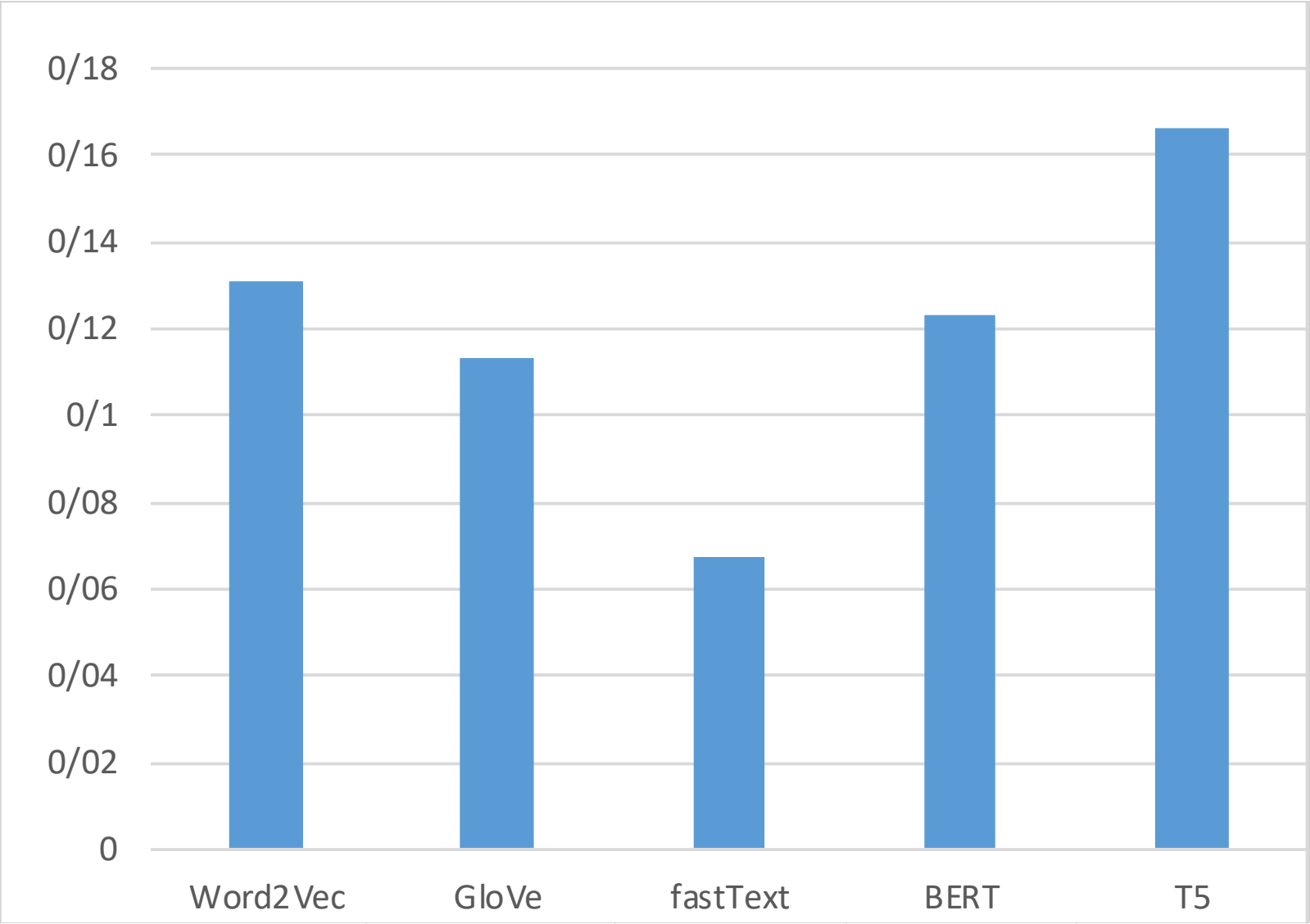

Figure 5: Final results for different embedding methods ($R_E$) calculated by eq. (5). (Averaged silhouette score.)

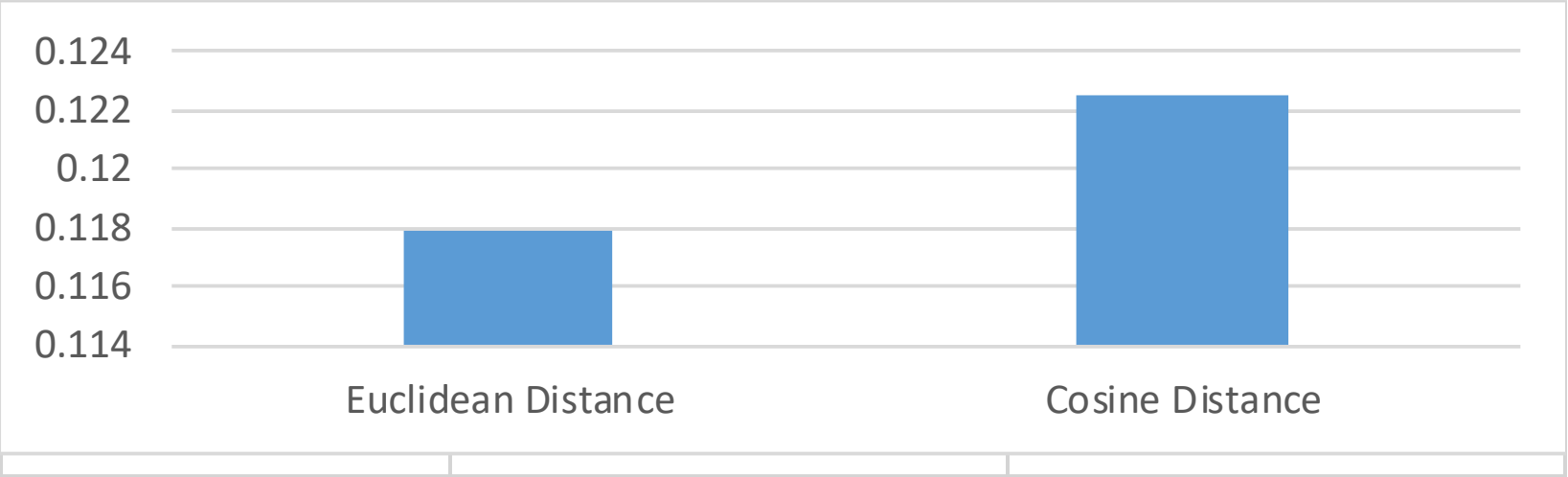

Figure 6: Final results for different distance metrics ($R_D$) obtained by eq. (6). (Averaged silhouette score.)

Table 4: Ranks of different embedding methods based on Table 3.

| | Euclidian Distance | | | | | Cosine Distance | | | | |
|---|---|---|---|---|---|---|---|---|---|---|
| | k-means | DBSCAN | OPTICS | spectral | Jarvis-Patrick | k-means | DBSCAN | OPTICS | spectral | Jarvis-Patrick |
| **Word2Vec** | 4 | 4 | 1 | 2 | 3 | 4 | 4 | 1 | 3 | 3 |
| **GloVe** | 2 | 3 | 3 | 4 | 4 | 2 | 1 | 2 | 4 | 4 |
| **fastText** | 5 | 2 | 5 | 5 | 5 | 3 | 3 | 5 | 5 | 5 |
| **BERT** | 3 | 5 | 2 | 1 | 2 | 5 | 2 | 4 | 2 | 2 |
| **T5** | 1 | 1 | 4 | 3 | 1 | 1 | 5 | 3 | 1 | 1 |

Table 5: Rank of cosine distance in comparison with Euclidian distance for different embedding and clustering methods.

| | **k-means** | **DBSCAN** | **OPTICS** | **spectral** | **Jarvis-Patrick** |
|---|---|---|---|---|---|
| **Word2Vec** | 1 | 2 | 2 | 2 | 1 |
| **Glove** | 1 | 1 | 1 | 2 | 1 |
| **Fast Text** | 1 | 2 | 2 | 2 | 1 |
| **Bert** | 1 | 1 | 2 | 1 | 1 |
| **T5** | 1 | 2 | 1 | 1 | 2 |

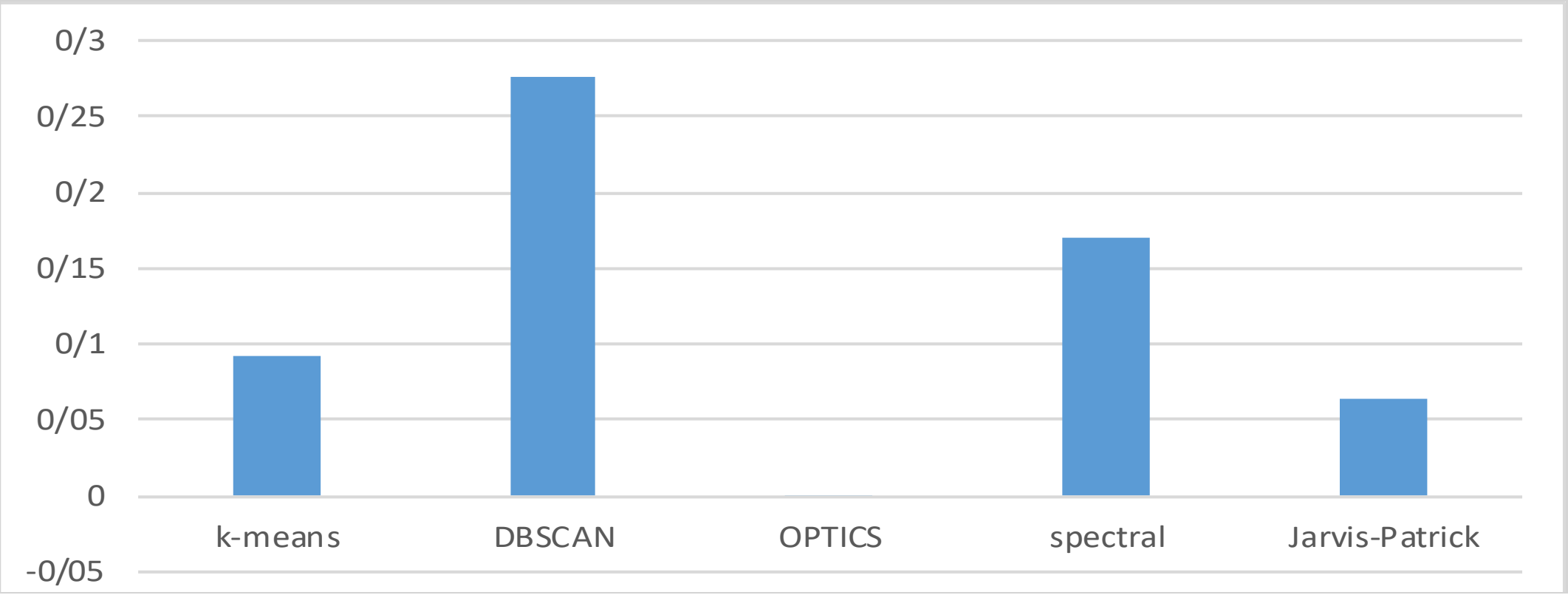

Figure 7: Final results for different clustering methods ($R_C$) calculated by eq. (7). (Averaged silhouette score.)

## 6- Conclusion

This paper has a comparative approach. This comparison is performed on topic detection. The COVID-19 dataset is also selected for this research. There are several approaches for topic detection from which the clustering approach is chosen in this paper. Clustering requires distance, and distance calculation needs embedding. Therefore, three goals are considered: performance evaluation of 1) embedding methods, 2) distance metrics, and 3) clustering methods. One of the advantages of this work is simultaneously investigating the three factors of embedding methods, distance metrics and clustering methods as well as their interaction. This research has three major questions: 1) Which embedding method has better performance in topic detection on COVID-19 tweets? 2) Which distance metric performs better in topic detection on COVID-19 tweets? 3) Which clustering method outperforms in topic detection on COVID-19 tweets?

Among the embedding methods, five methods are selected: Word2Vec, fastText, GloVe, BERT and T5, including earlier to new methods. Five methods of k-means, DBSCAN, OPTICS, spectral and Jarvis-Patrick, are investigated as clustering methods. Euclidian distance and cosine distance are also studied as the most important distance metrics for topic detection. First, parameter tuning experiments are performed, including more than 7500 cases. Then, all combinations of embedding methods with distance metrics and clustering methods with silhouette score are investigated. The number of these combinations consists of 50 cases. At first, the results of these 50 tests are studied. Then, the rank of each method is considered in all the experiments. At last, the independent variables of the research (embedding methods, distance metrics and clustering methods) are studied separately. In this case, the averaging is applied to neutralize the effect of control variables.

The experimental results show that T5 outperforms other embedding methods in terms of silhouette metric. In addition, T5 has the first rank in most cases when the clustering methods and distance metrics are changed. Therefore, it can be concluded that T5 is strongly better than other embedding methods. Word2Vec is in the second rank after T5. fastText has the weakest results since it has the last rank in most cases while the clustering methods and distance metrics are changed. Comparing distance metrics, cosine distance is weakly better. The cosine distance has better performance when k-means is used. This is also slightly true for the Jarvis-Patrick clustering method. Analyzing clustering methods shows that DBSCAN is superior to the other clustering methods.

### Decleration

The authors declare no conflict of interest in this study.

## 7- References